\documentclass[runningheads]{llncs}
\usepackage{graphicx}
\usepackage{comment}
\usepackage{amsmath,amssymb} 
\usepackage{bm}
\usepackage{color}

\usepackage{multirow}
\usepackage{algorithm}
\usepackage{algpseudocode}

\usepackage{footnote}
\makesavenoteenv{tabular}
\makesavenoteenv{table}

\usepackage{wrapfig,booktabs}
\usepackage{floatflt}

\usepackage{xr}
\makeatletter
\newcommand*{\addFileDependency}[1]{
  \typeout{(#1)}
  \@addtofilelist{#1}
  \IfFileExists{#1}{}{\typeout{No file #1.}}
}
\makeatother

\usepackage{footmisc}
\DefineFNsymbols{mySymbols}{{\ensuremath\ast}{\ensuremath\dagger}
{\ensuremath\ddagger}{\ensuremath\mathsection}{\ensuremath\mathparagraph}
{\ensuremath\parallel}{**}{\ensuremath{\dagger\dagger}}{\ensuremath{\ddagger\ddagger}}}
\setfnsymbol{mySymbols}

\newcommand{\tabincell}[2]{\begin{tabular}{@{}#1@{}}#2\end{tabular}}

\begin{document}

\pagestyle{headings}
\mainmatter
\def\ECCVSubNumber{3891}  
\title{Training Interpretable Convolutional Neural Networks by Differentiating Class-specific Filters}

\titlerunning{CSG CNN}
\author{
	Haoyu Liang\thanks{Haoyu Liang and Zhihao Ouyang contributed equally.}$^{1}$ \and
	Zhihao Ouyang$^{\ast 2,4}$ \and
	Yuyuan Zeng$^{2,3}$ \and
	Hang Su$^{\dagger 1}$ \and
	Zihao He$^{5}$ \and
	Shu-Tao Xia$^{2,3}$\and
	Jun Zhu\thanks{Hang Su and Jun Zhu are corresponding authors.}$^{1}$ \and
	Bo Zhang$^{1}$
}
\authorrunning{H. Liang et al.}
\institute{
$^1$Dept. of Comp. Sci. and Tech., BNRist Center, Inst. for AI, THBI Lab,\\
Tsinghua University, Beijing 100084, China\ $^2$Tsinghua SIGS, Shenzhen 518055, China\\
$^3$Peng Cheng Lab\ $^4$ByteDance AI Lab\ $^5$Dept. of CS, University of Southern California\\
\email{
\{lianghy18@mails, oyzh18@mails, zengyy19@mails, suhangss@mail,
xiast@sz, dcszj@mail, dcszb@mail\}.tsinghua.edu.cn\quad\quad
zihaoh@usc.edu}
}
\maketitle

\begin{abstract}
Convolutional neural networks (CNNs) have been successfully used in a range of tasks. However, CNNs are often viewed as ``black-box'' and lack of interpretability.
One main reason is due to the \textit{filter-class entanglement} -- an intricate many-to-many correspondence between filters and classes.
Most existing works attempt post-hoc interpretation on a pre-trained model, while neglecting to reduce the entanglement underlying the model.
In contrast, we focus on alleviating filter-class entanglement during training.
Inspired by cellular differentiation, we propose a novel strategy to train interpretable CNNs by encouraging \textit{class-specific filters}, among which each filter responds to only one (or few) class.
Concretely, we design a learnable sparse Class-Specific Gate (CSG) structure to assign each filter with one (or few) class in a flexible way.
The gate allows a filter's activation to pass only when the input samples come from the specific class.
Extensive experiments demonstrate the fabulous performance of our method in generating a sparse and highly class-related representation of the input, which leads to stronger interpretability.
Moreover, comparing with the standard training strategy, our model displays benefits in applications like object localization and adversarial sample detection.
Code link: \url{ https://github.com/hyliang96/CSGCNN}.
\end{abstract}

\section{Introduction}
Convolutional Neural Networks (CNNs) demonstrate extraordinary performance in various visual tasks~\cite{krizhevsky2012imagenet,he2016deep,girshick2015fast,he2017mask}.
However, the strong expressive power of CNNs is still far from being interpretable, which significantly limits their applications that require humans' trust or interaction, e.g., self-driving cars and medical image analysis~\cite{caruana2015intelligible,bojarski2017explaining}.

\begin{figure}[t]
   \centering
	  \includegraphics[width=0.9\linewidth]{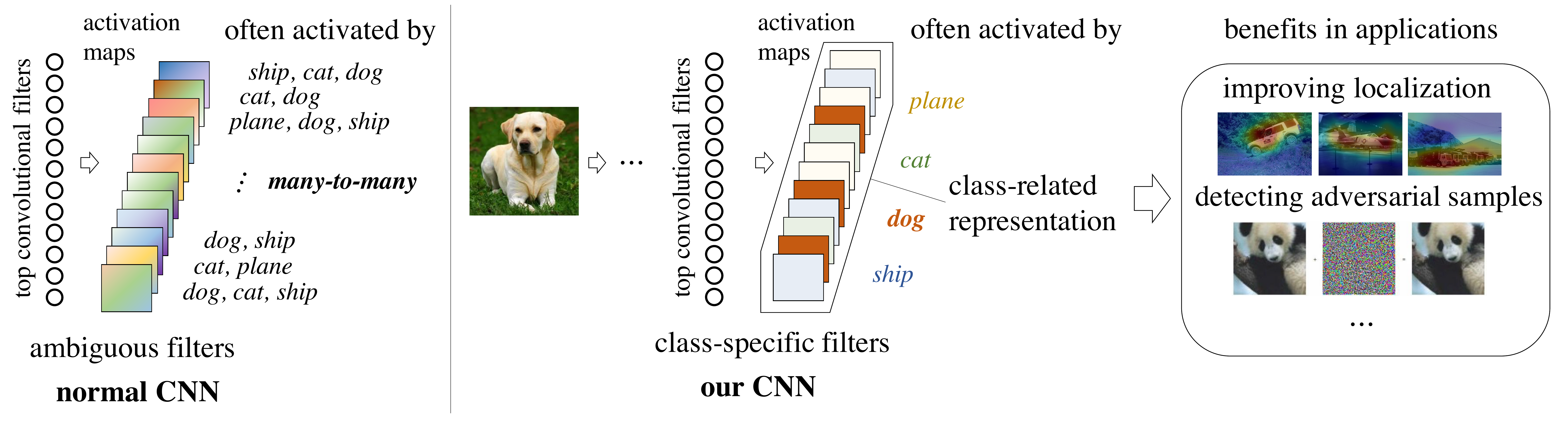}
	\caption{
	The motivation of learning class-specific filters.
	In a normal CNN, each filter responds to multiple classes, since it extracts a mixture of features from many classes~\cite{zhang2018interpretable}, which is a symptom of filter-class entanglement.
	In contrast, we enforce each filter to respond to one (or few) class, namely to be class-specific.
	It brings better interpretability and class-related feature representation.
	Such features not only facilitate understanding the inner logic of CNNs, but also benefits applications like object localization and adversarial sample detection.
	}
	\label{fig:intro_motivation}
\end{figure}

In this paper, we argue that filter-class entanglement is one of the most critical reasons that hamper the interpretability of CNNs.
The intricate \textit{many-to-many correspondence} relationship between filters and classes is so-called \textit{filter-class entanglement} as shown on the left of Fig.\ref{fig:intro_motivation}.
As a matter of fact, previous studies have shown that filters in CNNs generally extract features of a mixture of various semantic concepts, including the classes of objects, parts, scenes, textures, materials, and colors~\cite{zhang2018interpretable,bau2017network}.
Therefore alleviating the entanglement is crucial for humans from interpreting the concepts of a filter~\cite{zhang2018interpretable}, which has been shown as an essential role in the visualization and analysis of networks~\cite{olah2018the} in human-machine collaborative systems~\cite{zhang2017mining,zhang2017interactively}.
To alleviate the entanglement, this paper aims to learn \textit{class-specific filter} which responds to only one (or few) class.

Usually, it is non-trivial to deal with the entanglement, as many existing works show.
(1) Most interpretability-related research simply focuses on post-hoc interpretation of filters~\cite{bau2017network,szegedy2013intriguing}, which manages to interpret the main semantic concepts captured by a filter.
	However, post-hoc interpretation fails to alleviate the filter-class entanglement prevalent in pre-trained models.
(2) Many VAEs' variants ~\cite{higgins2017beta,burgess2018understanding,kim2018disentangling,chen2018isolating,kumar2017variational} and InfoGAN ~\cite{chen2016infogan} try to learn disentangled data representation with better interpretability in an unsupervised way. However, they are challenged by \cite{locatello2018challenging}, which proves that it's impossible unsupervised to learn disentangled features without proper inductive bias.

Despite the challenges above, it's reasonable and feasible to learning class-specific filters in \textit{high convolutional layers} in image classification tasks.
(1) It has been demonstrated that high-layer convolutional filters extract high-level semantic features which might relate to certain classes to some extent~\cite{zeiler2014visualizing};
(2) the redundant overlap between the features extracted by different filters makes it possible to learn specialized filters~\cite{prakash2019repr};
(3) specialized filters demonstrate higher interpretability~\cite{zhang2018interpretable} and better performance~\cite{prakash2019repr} in computer vision tasks;
(4) ~\cite{jiang2017learning,wang2018learning,martinez2019action} successfully learn class-specific filters in high convolution layers, though, under an inflexible predefined filter-class correspondence.

Therefore, we propose to learn class-specific filters in the last convolutional layer during training, which is inspired by cellular differentiation~\cite{smith1988inhibition}.
Through differentiation, the stem cells evolve to  functional cells with specialized instincts, so as to support sophisticated functions of the multi-cellular organism effectively.
For example, neural stem cells will differentiate into different categories like neurons, astrocytes, and oligodendrocytes through particular simulations from transmitting amplifying cells.
Similarly, we expect the filters in CNN to ``differentiate'' to disparate groups that have specialized responsibilities for specific tasks.
For specifically, we encourage the CNN to build a one-filter to one-class correspondence (differentiation) during training.

Specifically, we propose a novel training method to learn class-specific filters.
Different from existing works on class-specific filters that predefine filer-class correspondence, our model learns a \textit{flexible} correspondence that assigns only a necessary portion of filters to a class and allows classes to share filters.
Specifically, we design a learnable Class-Specific Gate (CSG) structure after the last convolutional filters,
	which assigns filters to classes and limits each filter's activation to pass only when its specific class(es) is input.
In our training process, we periodically insert CSG into the CNN and jointly minimize the classification cross-entropy and the sparsity of CSG, so as to keep the model's performance on classification meanwhile encourage class-specific filters.
Experimental results in Sec \ref{sec:exp} demonstrate that our training method makes data representation sparse and highly correlated with the labeled class, which not only illustrates the alleviation of filter-class entanglement but also enhances the interpretability from many aspects like filter orthogonality and filter redundancy.
Besides, in Sec \ref{sec:app} our method shows benefits in applications including improving objects localization and adversarial sample detection.

\textbf{Contributions\quad} The contributions of this work can be summarized as:
(1) we propose a novel training strategy for CNNs to learn a flexible class-filter correspondence where each filter extract features mainly from only one or few classes;
(2) we propose to evaluate filter-class correspondence with the mutual information between filter activation and prediction on classes, and moreover, we design a metric based on it to evaluate the overall filter-class entanglement in a network layer;
(3) we quantitively demonstrate the benefits of the class-specific filter in alleviating filter redundancy, enhancing interpretability and  applications like object localization and adversarial sample detection.

\section{Related Works}

Existing works related to our work include post-hoc filter interpretation, learning disentangled representation.

\textbf{Post-hoc Interpretation for Filters} is widely studied, which aims to interpret the patterns captured by filters in pre-trained CNNs.
Plenty of works visualize the pattern of a neuron as an image, which is the gradient~\cite{zeiler2014visualizing,mahendran2015understanding,simonyan2013deep} or accumulated gradient~\cite{mordvintsev2015inceptionism,olah2018the} of a certain score about the activation of the neuron.
Some works determine the main visual patterns extracted by a convolutional filter by treating it as a pattern detector~\cite{bau2017network} or appending an auxiliary detection module~\cite{gonzalez2018semantic}.
Some other works transfer the representation in CNN into an explanatory graph~\cite{zhang2017growing,zhang2018interpreting} or a decision tree~\cite{zhang2019interpreting,bai2019rectified}, which aims to figure out the visual patterns of filters and the relationship between co-activated patterns.
Post-hoc filter interpretation helps to understand the main patterns of a filter but makes no change to the existing filter-class entanglement of the pre-trained models, while our work aims to train interpretable models.

\textbf{Learning Disentangled Representation} refers to learning data representation that encodes different semantic information into different dimensions.
As a principle, it's proved impossible to learn disentangled representation without inductive bias~\cite{locatello2018challenging}.
Unsupervised methods such as variants of VAEs~\cite{kingma2014stochastic}  and InfoGAN~\cite{chen2016infogan} rely on regularization.
VAEs~\cite{kingma2014stochastic} are modified into many variants~\cite{higgins2017beta,burgess2018understanding,kim2018disentangling,chen2018isolating,kumar2017variational}, while their disentangling performance is sensitive to hyperparameters and random seeds.
Some other unsupervised methods rely on special network architectures including interpretable CNNs~\cite{zhang2018interpretable} and CapsNet~\cite{sabour2017dynamic}.
As for supervised methods,~\cite{thomas2018disentangling} propose to disentangle with interaction with the environment;~\cite{bouchacourt2018multi} apply weak supervision from grouping information, while our work applies weak supervision from classification labels.

\textbf{Class-specific Filters} has been applied in image and video classification task. The existing works focus on improve accuracy, including label consistent neuron~\cite{jiang2017learning} and filter bank ~\cite{wang2018learning,martinez2019action}. However, those works predefine an unlearnable correspondence between filters and classes where principally each filter responds to only one class and all classes occupy the same number of filters.
In contrast, this paper focuses on the interpretability of class-specific filters, and we propose a more flexible correspondence where similar classes can share filters and a class can occupy a learnable number of filters.
Therefore, our learnable correspondence helps reveal inter-class similarity and intra-class variability.

\section{Method}

Learning disentangled filters in CNNs alleviates filter-class entanglement and meanwhile narrows the gap between human concept and CNN's representations. In this section, we first present an ideal case of class-specific filters, which is a direction for our disentanglement training, and then we elaborate on our method about how to induce filter differentiation towards it in training an interpretable network.

\begin{figure}[t]
	\centering
	\includegraphics[width=0.9\textwidth]{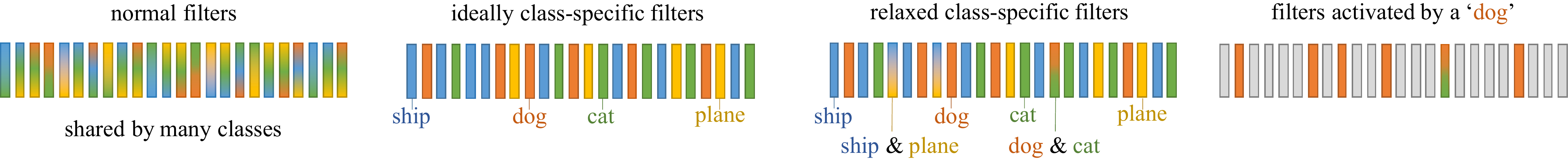}
\caption{
The intuition of learning class-specific filters.
In a standard CNN, a filter extracts a mixture of features from many classes~\cite{zhang2018interpretable}, which is a symptom of filter-class entanglement.
In contrast, an ideally class-specific filter extracts features mostly from only one class and a relaxed filter can be shared by few classes,
For flexibility, we actually apply the relaxed class-specific filter which is allowed to be shared by few classes.
Its activation to other classes is weak and has little effect on prediction.
}
\label{fig:intuition}
\end{figure}

\subsection{Ideally Class-Specific Filters}
This subsection introduces an ideal case for the target of our filter disentanglement training.
As shown in Fig.\ref{fig:intuition}, each filter mainly responds to (i.e. relates to) only one class.
We call such filters \textit{ideally class-specific} and call disentangling filter towards it in training as \textit{class-specific differentiation} of filters.

To give a rigorous definition of ``ideally class-specific'' for a convolutional layers, we use a matrix $G\in [0,1]^{C\times K}$ to measure the relevance between filters and classes, where $K$ is the number of filters, $C$ is the number of classes. Each element $G_c^k \in [0,1]$ represents the relevance between the $k$-th filter and $c$-th class, (a larger $G_c^k$ indicates a closer correlation).
As shown in Fig.\ref{fig:framework}, the $k$-th filter extracts features mainly in the $c$-th class iif $G_c^k = 1$.
Denote a sample in dataset $D$ as $(x,y)\in D$ where $x$ is an image and $y \in \{1,2,...,C\}$ is the label.
Given $(x,y)$ as an input, we can index a row $G_y\in [0,1]^{K}$ from the matrix $G$, which can be used as a gate multiplied to the activation maps to shut down those irrelevant channels.
Let $\tilde{{y}}$ be the probability vector predicted by the STD path, and $\tilde{{y}}^G$ be the probability vector predicted by the CSG path where the gate $G_y$ is multiplied on the activation maps from the penultimate layer.
Thus, we call convolutional filters as {\it ideally class-specific} filters, if there exists a $G$ (all columns $G_k$ are \textit{one-hot}) that raises little difference between the classification performance of $\tilde{{y}}^G$ and $\tilde{{y}}$.

\subsection{Problem formulation}\label{sec:form}
In order to train a CNN towards differentiating filters to class-specific meanwhile keep classification accuracy,  we introduce a Class-Specific Gate (CSG) path in addition to the standard (STD) path of forward propagation.
In the CSG path, channels are selectively blocked with the learnable gates.
This path's classification performance is regarded as a regularization for filter differentiation training.

To derive the formulation of training a CNN with class-specific filters, we start from an original problem that learns ideally class-specific filters and then relax the problem for the convenience of a practical solution.

\begin{figure}[t]
\begin{minipage}[b]{0.5\linewidth}
\includegraphics[width=1\textwidth]{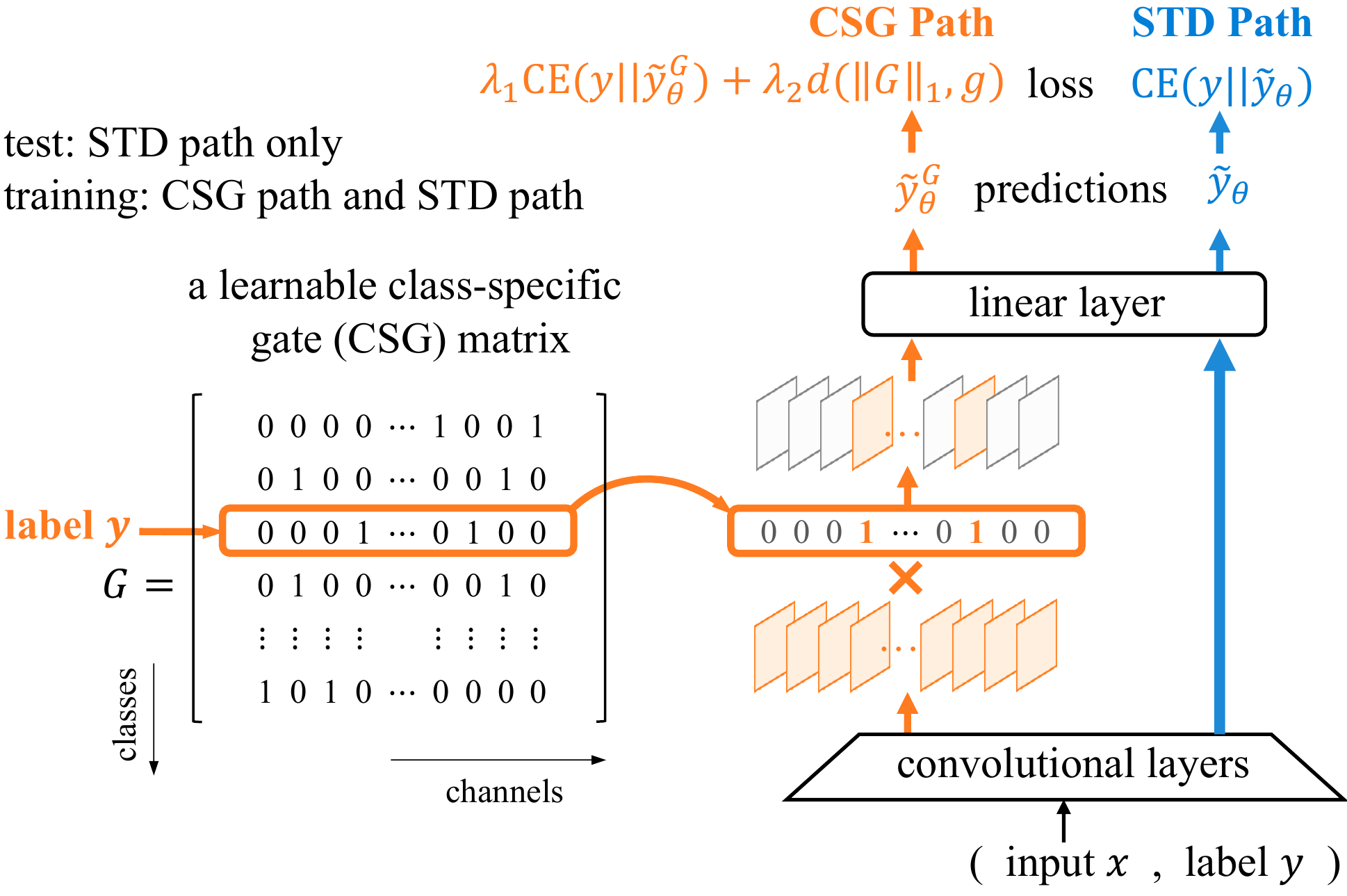}
\label{fig:framework}
\end{minipage}
\hfill
\begin{minipage}[b]{0.45\linewidth}
\caption{
In Class-Specific Gate (CSG) training, we \emph{alternately} train a CNN through the CSG path and the standard (STD) path.
In the CSG path, activations after the penultimate layer (i.e. the last convolution) pass through learnable gates indexed by the label.
In training, network parameters and the gates are optimized to minimize the cross-entropy joint with a sparsity regularization for the CSG matrix.
In testing, we just run the STD path.
}
\end{minipage}
\end{figure}

\textbf{The Original Problem} is to train a CNN with ideally class-specific filters. See Fig.\ref{fig:framework} for the network structure. The network with parameters $\theta$ forward propagates in two paths: (1) the standard (STD) path predicting $\tilde{{y}}_\theta$, and (2) the CSG path with gate matrix $G$ predicting $\tilde{{y}}^G_\theta$ where activations of the penultimate layer are multiplied by learnable gates $G_y$ for inputs with label $y$.

In order to find the gate matrix $G$ that precisely describes the relevance between filters and classes, we search in the binary space for a $G$ that yields the best classification performance through the CSG path, i.e., to solve the optimization problem $\Phi_0(\theta)= \min\limits_{G} \text{CE}(y||\tilde{{y}}^G_\theta)$\footnote{$\text{CE}(y||\tilde{{y}}^G_\theta)=- \frac{1}{|D|}\sum_{(x,y)\in D} \log( (\tilde{{y}}^G_\theta)_y )$, where $\tilde{{y}}^G_\theta$ is a predicted probability vector.}
	s.t. $\forall k\in\{1,2,..., K\}, G_k$ is one-hot. $\Phi_0$ evaluates the performance of the CNN with differentiated filters.
Therefore, it is natural to add $\Phi_0$ into training loss as a regularization that forces filters to be class-specific. Thus, we get the following formulation of the \textit{original problem} to train a CNN towards ideally class-specific filters as
\begin{align}
    \min_\theta L_0(\theta) = \text{CE}(y||\tilde{{y}}_\theta) + \lambda_1 \Phi_0(\theta). \label{equ:loss1}
\end{align}
where the ${CE}(y||\tilde{{y}}_\theta)$ ensures the accuracy and $\lambda_1 \Phi_0(\theta)$ encourage sparsity of $G$.

However, the original problem is difficult to solve in practice. On the one hand, the assumption that each filter is complete one-filter/one-class assumption hardly holds, since it is usual for several classes to share one high-level feature in CNNs; on the other hand, binary vectors in a non-continuous space are difficult to optimize with gradient descent.

\textbf{Relaxation\quad}To overcome the two difficulties in the original problem above, we relax relax the one-hot vector $G^k$ to a sparse continuous vector $G^k\in[0,1]^C$ where at least one element equals to 1, i.e., $\left\|G^k\right\|_\infty=1$.
To encourage the sparsity of $G$, we introduce a regularization $d(\left\| G\right\|_1,g)$ that encourages the L1 \textit{vector norm} $\left\| G\right\| _1$  not to exceed the upper bound $g$ when $\left\| G\right\|_1 \geq g$, and has no effect when $\left\| G\right\|_1 < g$.
A general form for $d$ is  $d(a,b)=\psi(\mathrm{ReLU}(a-b))$, where $\psi$ can be any norm, including L1, L2 and smooth-L1 norm.
Besides, we should set $g\geq K$ because $\left\|G\right\|_1 \geq K$ which is ensured by $\left\| G^k\right\| _\infty=1$.
Using the aforementioned relaxation, $\Phi_0$ is reformulated as
\begin{equation}\label{equ:phi}
    \Phi(\theta) = \min_{G} \{ \text{CE}(y||\tilde{y}^G_\theta)  +
     \mu d( \left\|G\right\|_1 ,g) \} \quad
     s.t.\  G \in V_G,
\end{equation}
where the set $V_G=\{G\in [0,1]^{C\times K} : \left\| G^k \right\|_\infty  = 1\}$ and $\mu$ is a coefficient to balance classification and sparsity. $\Phi$ can be regarded as a loss function for \textit{filter-class entanglement}, i.e., a CNN with higher class-specificity has a lower $\Phi$.

Replacing $\Phi_0$ in Eq.~\eqref{equ:loss1} with $\Phi$, we get
$ \min_{\theta} \text{CE}(y||\tilde{y}_\theta) + \lambda_1 \Phi(\theta)$ as an intermediate problem. It is mathematically equivalent if we move $\min_G$ within $\Phi$ to the leftmost and replace $\lambda_1\mu$ with $\lambda_2$. Thus, combining Eq.~\eqref{equ:loss1} \eqref{equ:phi}, we formulate a \textit{relaxed problem} as
\begin{eqnarray}\label{equ:relax}
\min_{\theta,G} L(\theta,G) =
	\text{CE}(y||\tilde{y}_\theta) +
	 \lambda_1  \text{CE}(y||\tilde{y}^G_\theta)  +
	 \lambda_2  d( \left\|G\right\|_1 ,g)\quad
	 s.t.\ G \in V_G.
\end{eqnarray}

The relaxed problem is easier to solve by jointly optimizing $\theta$ and $G$ with gradient, compared to either the discrete optimization in the original problem or the nested optimization in the intermediate problem. Solving the relaxed problem, we can obtain a CNN for classification with class-specific filters, where $G$ precisely describes the correlation between filters and classes.

\begin{algorithm}[t]
\caption{ CSG Training}
\label{alg:train}
\begin{algorithmic}[1]
\For{$e$ in epochs}
    \For{$n$ in batches}
        \If{ $e\ \% \ period \leq epoch\_num\_for\_CSG$ }
            \State $\tilde{y}^G_\theta \leftarrow$ prediction through the CSG path with $G$
            \State $\mathcal{L} \leftarrow
             	\lambda_1  \text{CE}(y||\tilde{y}^G_\theta)  +
	            \lambda_2  d( \left\|G\right\|_1 ,g) $
            \State $G \leftarrow G - \epsilon\frac{ \partial \mathcal{L} }{ \partial G}$
            	\Comment{update $G$ using the gradient decent}
            \State $G^k \leftarrow G^k / \left\| G^k \right\|_\infty$
            	\Comment{normalize each column of $G$}
            \State $G \leftarrow \text{clip}(G,0,1)$
        \Else
            \State $\tilde{y}_\theta \leftarrow$ prediction through the STD path
            \State $\mathcal{L} \leftarrow \text{CE}(y||\tilde{y}_\theta)$
         \EndIf
        \State $\theta \leftarrow \theta - \epsilon \frac{\partial \mathcal{L} }{ \partial \theta}$
        \EndFor
\EndFor
\end{algorithmic}
\end{algorithm}

\subsection{Optimization}\label{sec:opt}

To solve the optimization problem formulated in Eq.~\eqref{equ:relax} we apply an approximate projected gradient descent (PGD): when $G$ is updated with gradient, $G^k$ will be normalized by $\left\| G^k \right\|_\infty$ to ensure $ \left\| G^k \right\|_\infty=1$, and then clipped into the range $[0,1]$.

However, it is probably  difficult for the normal training scheme due to poor convergence.
In the normal scheme, we predict through both CSG and STD paths to directly calculate $L(\theta,G)$ and update $\theta$ and $G$ with gradients of it.
Due to that most channels are blocked in the CSG path, the gradient through the CSG path will be much weaker than that of STD path, which hinders converging to class-specific filters.

To address this issue, we propose an \textit{alternate training scheme} that the STD/CSG path works alternately in different epochs.
As shown in Algorithm \ref{alg:train}, in the epoch for CSG path, we update $G,\theta$ with the gradient of $\lambda_1  \text{CE}(y||\tilde{y}^G_\theta)  + \lambda_2  d( \left\|G\right\|_1 ,g) $  , and in the STD path we update $\theta$ with the gradient of  $\text{CE}(y||\tilde{y}_\theta)$
In this scheme, the classification performance fluctuates periodically at the beginning but the converged performance is slightly better than the normal scheme in our test. Meanwhile, the filters gradually differentiate into class-specific filters.

\section{Experiment}\label{sec:exp}
In this section, we conduct five experiments.
We first delve into CSG training from three aspects, so as to respectively study the effectiveness of CSG training, the class-specificity of filters and the correlation among class-specific filters.
Especially, to measure filter-class correspondences we apply the mutual information (MI) between each filter's activation and the prediction on each class.
In the following parts, we denote our training method Class-Specific Gate as \emph{CSG}, the standard training as \emph{STD}, and CNNs trained with them as \emph{CSG CNNs} and \emph{STD CNNs}, respectively.

\textbf{Training\quad} We use CSG/STD to train ResNet-18/20/152s~\cite{he2016deep} for classification task on  CIFAR-10~\cite{krizhevsky2009learning}/ImageNet~\cite{imagenet_cvpr09}/PASCAL VOC 2010~\cite{pascal-voc-2010} respectively. We select six animals from PASCAL VOC and preprocess it to be a classification dataset. The ResNet-18/20s are trained from scratch and the ResNet-152s are finetuned from ImaageNet. See Appendix~\ref{app:train} for detailed training settings.

\subsection{Effectiveness of CSG Training}\label{sec:train}

First of all, we conduct experiments to verify the effectiveness of our CSG training in learning a sparse gate matrix and achieve high class-specificity of filters.

\begin{table}[t]
\label{tab:metric for CSG}
\begin{scriptsize}
\begin{center}
\caption{Metrics of the STD CNN (baseline) and the CSG CNN (Ours).}
\begin{tabular}{l|l|l|l|l|c|c|c|c}
\toprule
 Dataset & Model & $C$ & $K$ & Training  & Accuracy & MIS & L1-density & L1-interval \footnote{\label{note:ub}
	They are $[\frac{1}{C}, \frac{g}{CK}]$ -- the theoretical convergence interval for L1-density of CSG CNNs,
	where $g$ is the upper bound for $\left\|G\right\|_1$ in Eq.~\eqref{equ:phi}. See Appendix~\ref{app:l1density} for the derivation.
	When the dataset has numerous classes like ImageNet, the L1-density can drop much lower than $\frac{g}{CK}$ due to the projection in our approximate PGD.} \\
\midrule
 	\multirow{2}*{CIFAR-10} & \multirow{2}*{ResNet20} &
 	\multirow{2}*{10} & \multirow{2}*{64} &
 	 CSG &
	\textbf{0.9192}  &  \textbf{0.1603}     & 0.1000 	& $[0.1,0.1]$  	\\
	~ & ~& ~& ~& STD &
	0.9169 &	 0.1119  & - & - \\
\midrule
	\multirow{2}*{ImageNet}  & \multirow{2}*{ResNet18} &
	\multirow{2}*{1000} & \multirow{2}*{512} &
	 CSG  & 0.6784 & \textbf{0.6259} & 0.0016  & $[0.001,0.1]$ \\
	~ & ~& ~& ~& STD & \textbf{0.6976}  & 0.5514 & - & - \\
\midrule
	\multirow{2}*{\tabincell{c}{PASCAL\\VOC 2010}} & \multirow{2}*{ResNet152} &
	\multirow{2}*{6} & \multirow{2}*{2048} &
	CSG &\textbf{0.8518}&\textbf{0.1998}& 0.1996 & $[0.1667,0.2]$\\
	~ & ~ & ~&~&STD  &  0.8447 &  0.1428 &  - & -  \\
\bottomrule
\end{tabular}
\end{center}
\end{scriptsize}
\end{table}

\textbf{Quantitative Evaluation Metrics\quad}
To evaluate the effectiveness of CSG, we calculate 3 metrics:
L1-density, mutual information score, and classification accuracy.
(1) Accuracy measures the classification performance.
(2) To measure the correspondence between filters and classes, we propose the mutual information (MI) matrix $M\in \mathbb{R}^{K\times C}$ where $M_{kc} = \mathrm{MI}(a_k||\mathbf{1}_{y=c})$ is the MI between $a_k$ -- the activation of filter-$k$ and class-$c$. To calculate the MI, we sample $(x, y)$ across the dataset, $a_k$ (the globally avg-pooled activation map of filter $k$ over all the sampled $x$) is a continuous variable, and $\mathbf{1}_{y=c}$ is a categorical variable. The estimation method ~\cite{ross2014mutual} for the MI between them is implemented in the API `sklearn.feature\_selection.mutual\_info\_classif'.
Base on this, we propose a mutual information score $MIS=\mathrm{mean}_k \max_c M_{kc}$ as an overall metric of class-specificity of all filters. Higher MIS indicates higher class-specificity, aka, lower filter-class entanglement.
(3) The L1-density $=\frac{\left\| G\right\|_1}{KC}$ is the L1-norm of CSG normalized by the number of elements, which measures the sparsity of CSG.

Table \ref{tab:metric for CSG} shows that CSG CNNs are comparable to or even slightly outperforms STD CNNs in test accuracy, while the CSG CNNs have MIS much higher than STD CNNs and the L1-density of $G$ is limited in its theoretical convergence interval.
These metrics quantitatively demonstrate CSG's effectiveness on learning a sparse gate matrix and class-specific filters without sacrifice on classification accuracy.

\begin{floatingfigure}[r]{0.45\textwidth}
	\centering
    \includegraphics[width=0.45\textwidth]{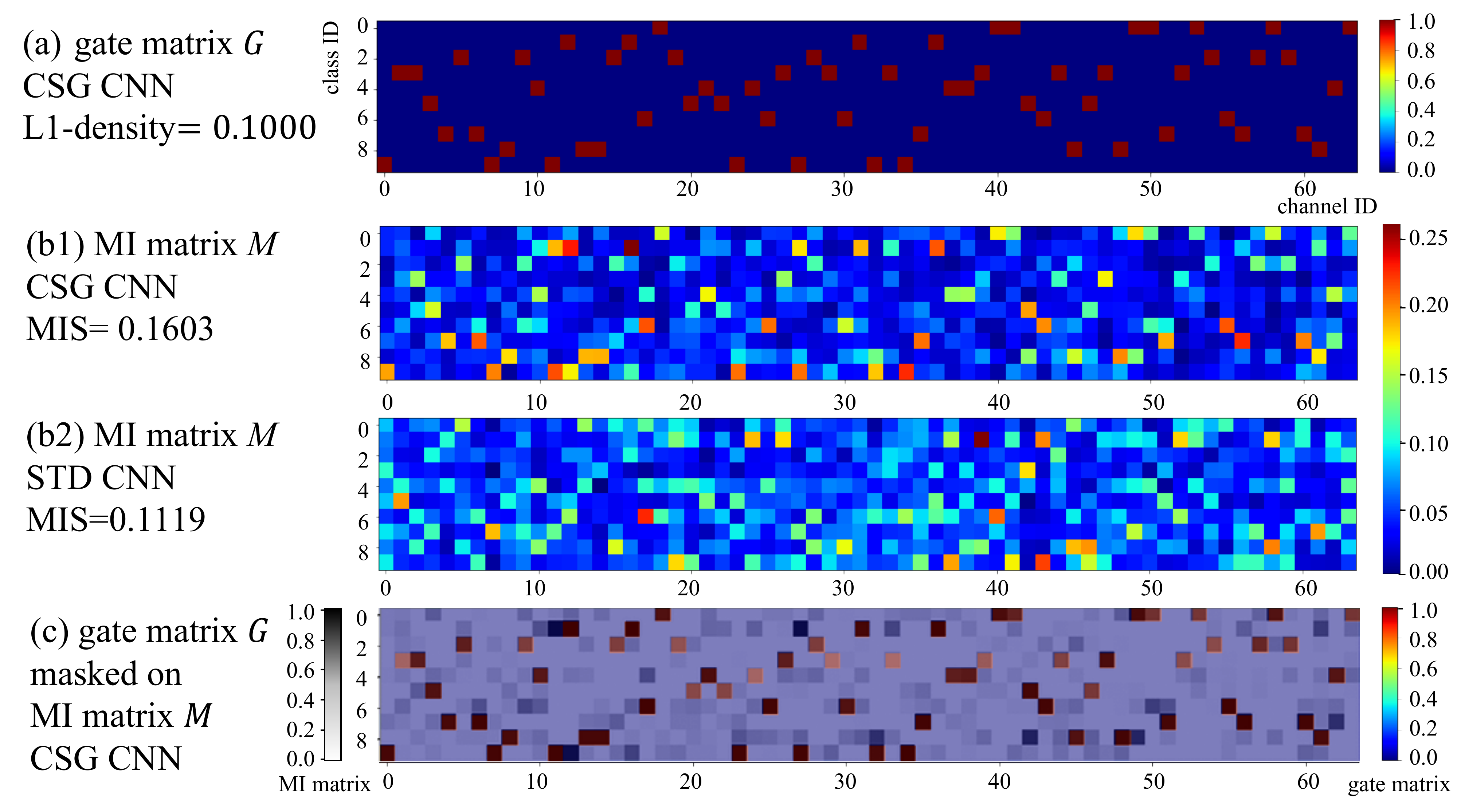}
	\caption{
		(a) visualizes the CSG matrix of CSG CNN to verify its sparsity.
		(b1,b2) compare the MI matrices of CSG/STD CNN and their MIS.
		(c) is got by overlapping (a) and (b1).
    }
    \label{fig:CD_Gates}
\end{floatingfigure}

\textbf{Visualizing the Gate/MI Matrices\quad}
To demonstrate that the relevance between the learned filters and classes is exactly described by the gate matrix, we visualize gate matrix and MI matrices in Fig.\ref{fig:CD_Gates}.
(a) demonstrates that CSG training yields a sparse CSG matrix where each filter is only related to one or few classes.
(b1,b2) shows that CSG CNN has sparser MI matrices and larger MIS compared to STD CNN.
(c) shows the strongest elements in the two matrices almost overlaps, which indicates that CSG effectively learns filters following the guidance of the CSG matrix.
These together verify that CSG training effectively learns a sparse gate matrix and filters focusing on the one or few classes described by the gates.

\subsection{Study on Class-Specificity}\label{sec:filter}

In this subsection, we study the property of class-specific filters and the mechanism of filter differentiation through experiments on ResNet20 trained in \ref{sec:train}.

\begin{figure}[t]
\centering
\includegraphics[width=0.8\textwidth]{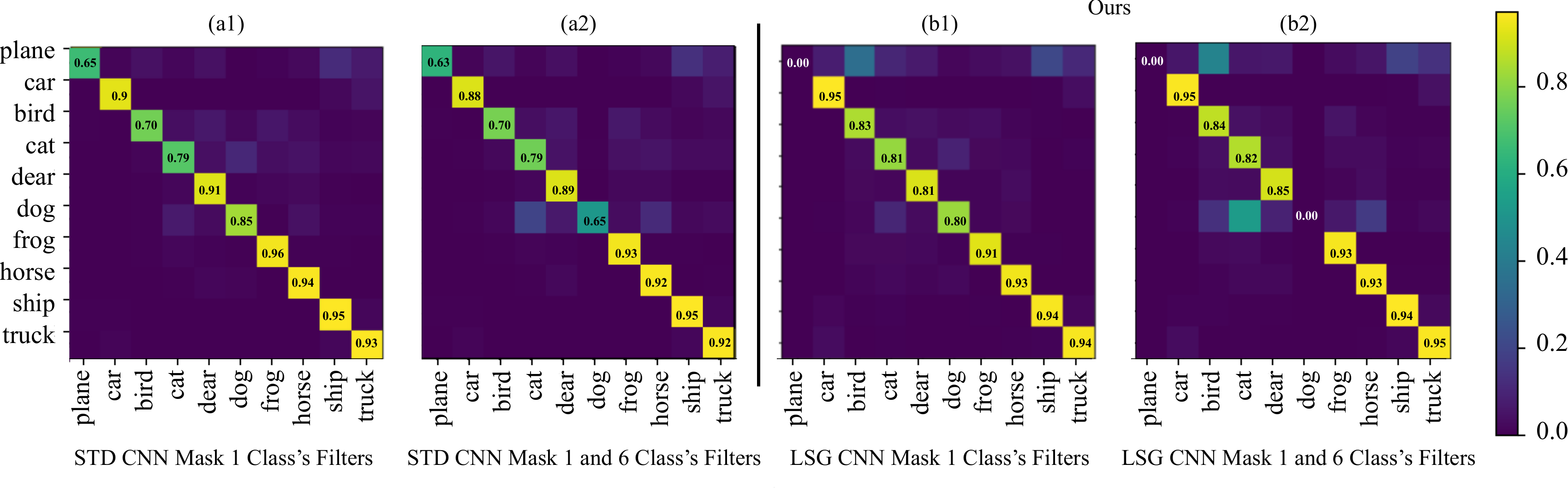}
    \caption{Classification confusion matrix for STD / CSG models when masking filters highly related to the first class (a1, b1), and the first and sixth classes (a2, b2).
     }
    \label{fig:mask_CD}
\end{figure}

\textbf{Indispensability for Related Class\quad}
It is already shown by the MI matrix that filters are class-specific, while we  further reveal that the filters related to a class are also {\it indispensable} in recognizing the class.
We remove the filters highly related to certain class(es) referencing the gate matrix (i.e., $G_c^k>0.5$), and then visualize the classification confusion matrices.
For STD CNN, we simply remove 10\% filters that have the largest average activation to certain classes across the dataset.
As shown in Fig.\ref{fig:mask_CD}, when filters highly related to ``plane'' are removed, the CSG CNN fails to recognize the first class ``plane''; nevertheless, the STD CNN still manages to recognize ``plane''.
Analogously, when removed the filters highly related to the first and sixth classes, we observe a similar phenomenon.
This demonstrates that the filters specific to a class are indispensable in recognizing this class. Such a phenomenon is because the filters beyond the group mainly respond to other classes and hence can't substitute for those filters.

\begin{floatingfigure}[r]{0.55\textwidth}
	\centering
	\includegraphics[width=0.5\textwidth]{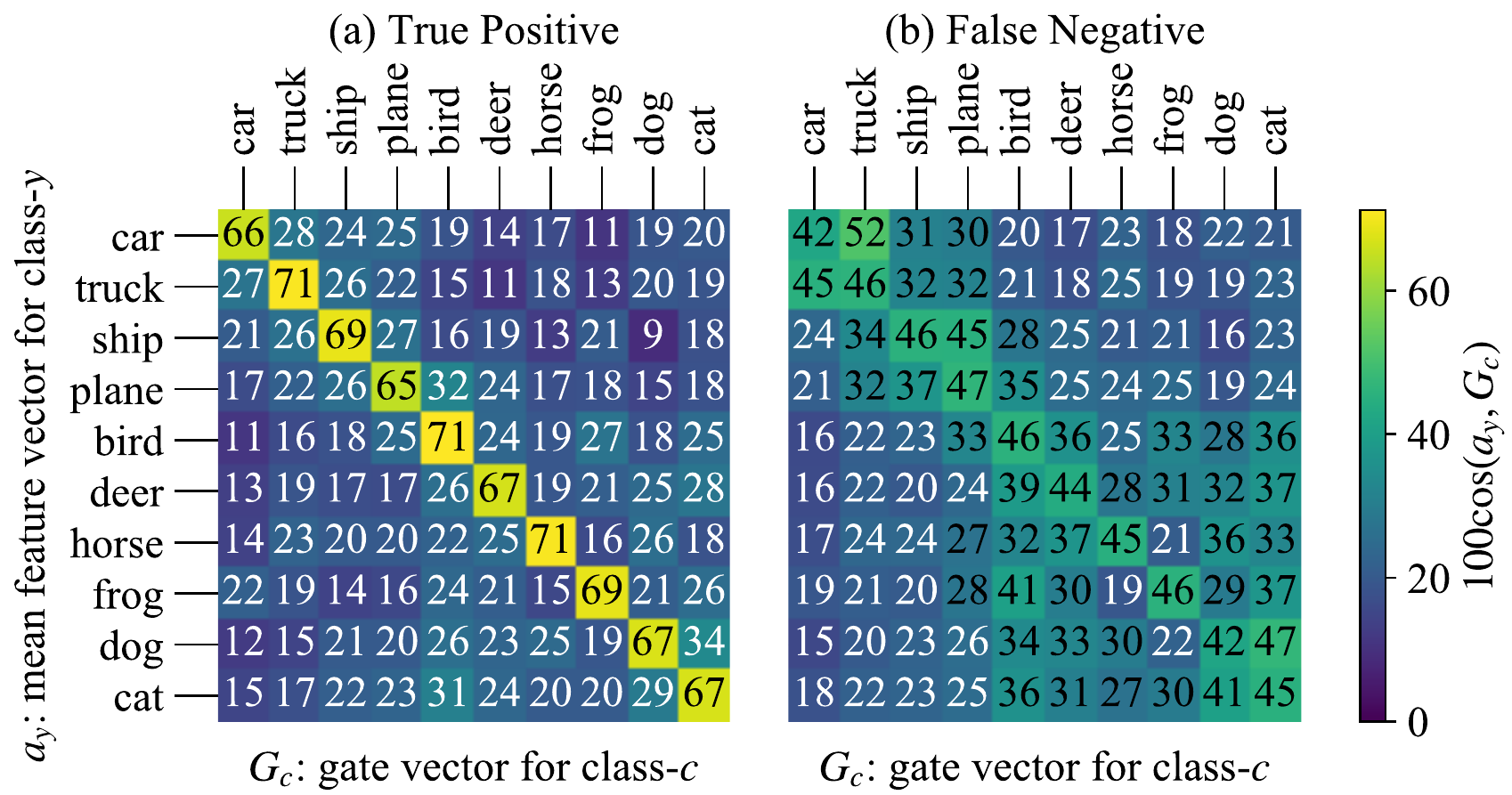}
	\caption{
		The cosine similarity between mean feature vectors for a class (y-axis) and a row in the CSG matrix (x-axis), calculated on all TP/FN samples.
		We reorder classes in CIFAR10 for better visualization.
	}
	\label{fig:corr_fw}
\end{floatingfigure}

\textbf{Mechanism of Class-Specific Differentiation\quad}\label{sec:sim}
In this part we reveal that the mechanism of filter differentiation by studying the directional similarity between features and gates, and by the way, explain misclassification with the CSG.
We use cosine to measure the directional similarity $S_y^c$ between $a_y\in\mathbb{R}^K$ the mean feature (average-pooled feature from class-specific filters) for class $y$, and $G_c$  the $c$-th row of the CSG matrix (see Appendix~\ref{app:sim} for details).
We calculate $S_y^c$ over all true positive (TP) and false negative (FN) images and obtain similarity matrices $S_{TP},S_{FN}\in [0,1]^{C\times C}$ respectively, as shown in Fig.\ref{fig:corr_fw}.

From the figure, we observe two phenomena and provide the following analysis.
(1)
TP similarity matrix is diagonally dominant.
This reveals mechanism of learning class-specifics: CSG forces filters to yield feature vectors whose direction approaches that of the gate vector for its related class.
(2)
FN similarity matrix is far from diagonally dominant and two classes with many shared features, such as car \& truck and ship \& plane, have high similarity in the FN similarity matrix.
CSG enlightens us that hard samples with feature across classes tend to be misclassified.
Thus, the mechanism of misclassification in the CSG CNN is probably that the features across classes are extracted by the shared filters. To some extent, it proves differentiation is beneficial for accuracy.

\subsection{Correlation Between Filters} \label{sec:filter_corr}
We further designed several experiments to explore what happen to filters (why they are class-specified). Our analysis shows inter-class filters are approximately orthogonal and less redundant, and the class-specific filters yield highly class-related representation.

\textbf{Fixing the Gates\quad}\label{sec:tight}
To make it convenient to study inter-class filters, we group filters in a tidier way --  each class monopolizes $m_1$ filters and $m_2$ extra filters are shared by all classes.
 This setting can be regarded as tightening the constraint on the gate matrix to $G\in \{0,1\}^{C\times K}$, according to \ref{sec:opt} (see Appendix~\ref{app:tight} for illustration).
The corresponding CSG matrix is fixed during training.
We train an AlexNet~\cite{krizhevsky2012imagenet} ($m_1=25, m_2=6$) and a ResNet20 ($m_1=6, m_2=4$) by STD and by CSG in this way.
Models trained with this setting naturally inherits all features of previous CSG models and has tidier filter groups.

\textbf{Filter Orthogonality Analysis \quad}\label{sec:amb}

\begin{figure}[t]
	\centering
    \includegraphics[width=0.9\textwidth]{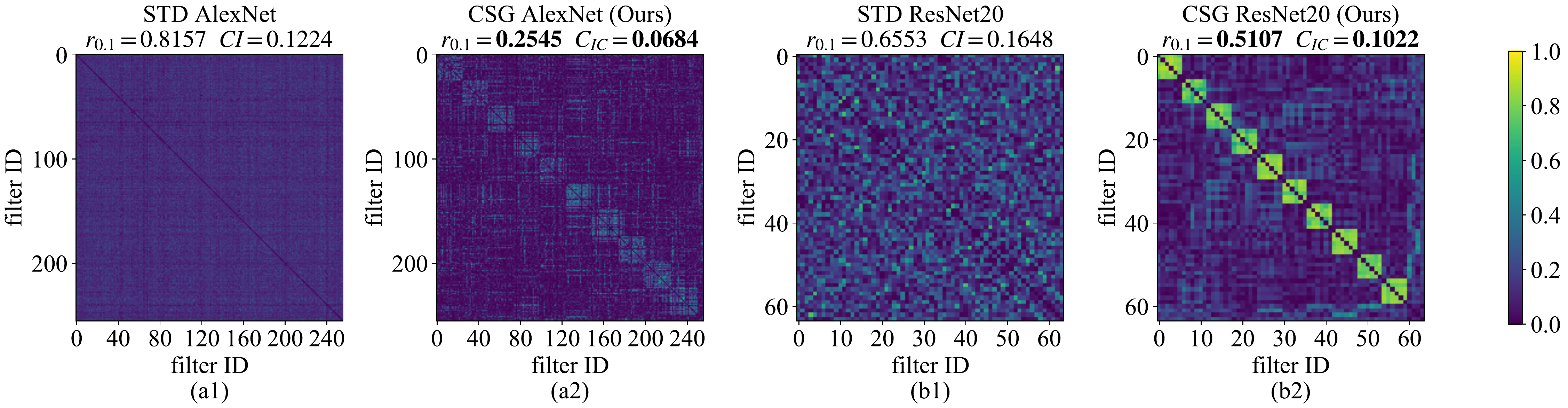}
    \caption{The correlation matrix of filters (cosine between filters' weights) in AlexNet and ResNet20 trained with STD/CSG. $r_{0.1}$: the ratio of elements$\geq 0.1$, measures filter redundancy. $C_{IC}$: the inter-class filter correlation, measures inter-class filter similarity.}
    \label{fig:correlation}
\end{figure}

To study the orthogonality between filters, we
evaluate filter correlation with the cosine of filters' weights.
The correlations between all filters are visualized as correlations matrices $\mathcal{C}$ in Fig.\ref{fig:correlation}.
In subfigures (a1, b1) for STD models, the filters are randomly correlated with each other.
In contrast, in subfigures (a2, b2) for CSG models, the matrices are approximately block-diagonal, which means the correlation between the filters is limited to several class-specific filter groups.
This indicates that filters for the same class are highly correlated (non-orthogonal) due to the co-occurrence of features extracted by them,
while filters for different classes are almost uncorrelated (orthogonal) for the lack of co-occurrence.
See Appendix~\ref{app:orthogonnal} for a detailed explanation for orthogonality.

\begin{floatingfigure}[r]{0.35\textwidth}
	\centering
      \includegraphics[width=0.35\textwidth]{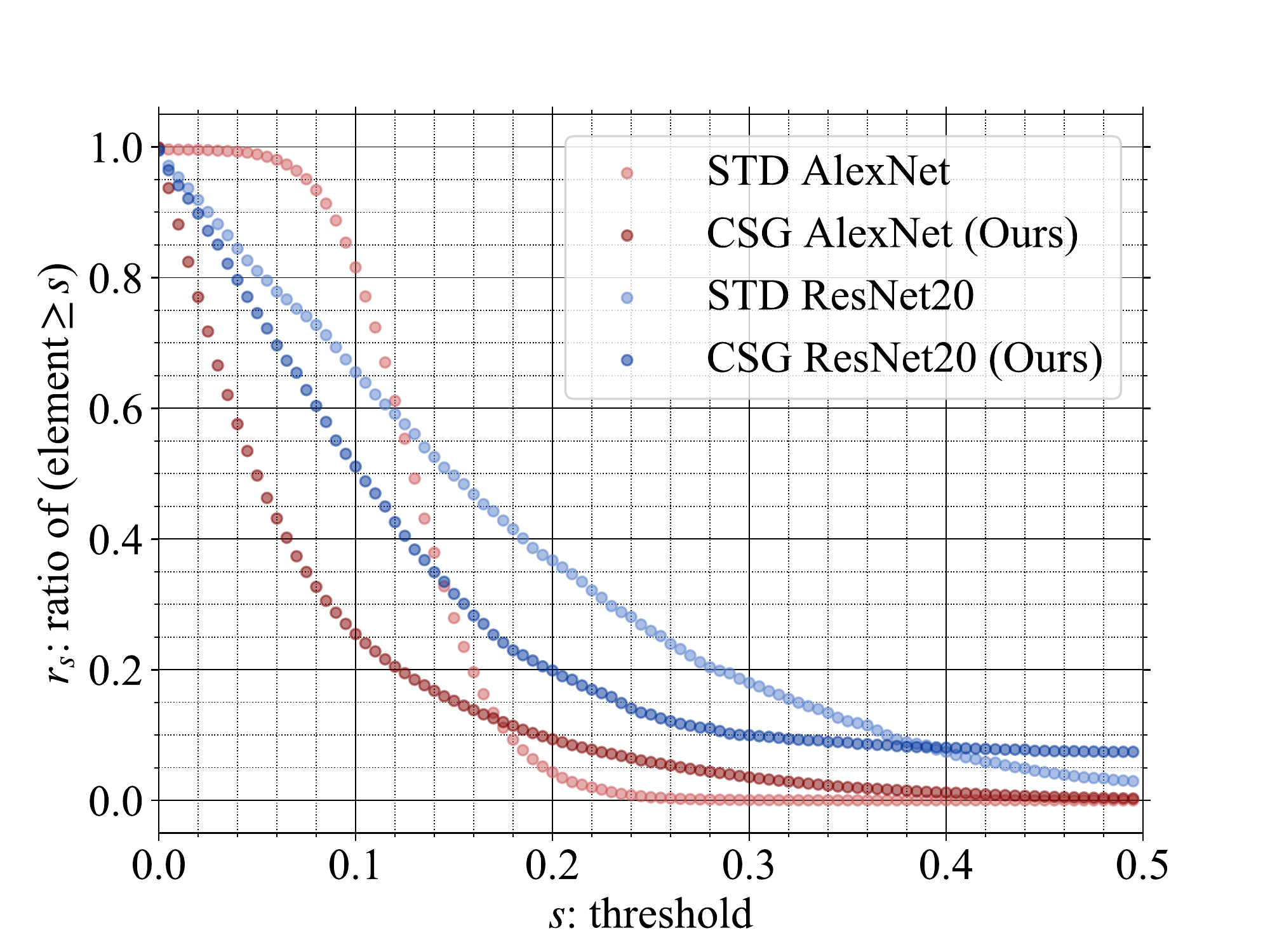}
	  \caption{the ratio of elements larger than a varying threshold in correlation matrices.}
	  \label{fig:correlation-curve}
\end{floatingfigure}

\textbf{Filter Redundancy \quad} To verify that the redundancy of filters are reduced by CSG, we research the inter-class filter correlation.
Let the filter-$i,j$  are correlated if their correlation $\mathcal{C}_{i,j}\geq s$ ($s$ is a threshold). We calculate $r_s$ the ratio of such elements in a correlation matrix and plot the results in Fig.\ref{fig:correlation-curve}.
It shows that CSG significantly reduces the redundant correlation ($\mathcal{C}_{i,j}<s=0.1$) between most filter pairs from different groups. We explain the reduction of filter redundancy as a natural consequence of encouraging inter-class filters to be orthogonal. For a set of filter groups orthogonal to each other, a filter in any group can not be a linear representation with the filers from other groups. This directly avoids redundant filter across groups. Besides, experiments in~\cite{prakash2019repr} also verify the opinion that filter orthogonality reduces filter redundancy.

\textbf{Highly Class-Related Representation\quad}\label{sec:label_related}

Based on filter correlation, we further find that filters trained with CSG yield highly class-related representation,
namely the representation of an image tends to exactly correspond to its labeled class rather than other classes.
Because the implied class is mainly decided on which filters are most activated, meanwhile those filters are less activated by other classes and less correlated to the filters for other classes.

To verify this reasoning, we analyze the correlation between the filters highly activated by each class.
First, we pick out $m_1$ filters that have the strongest activation to class $c$, denoted as group $A_c$.
We define inter-class filter correlation as the average correlation between filters in different classes' group $A_c$: $ C_{\text{IC}} = \frac{1}{C(C-1)m^2}$ $\sum_{c}$ $\sum_{c\neq c'}$ $\sum_{k\in A_c}$ $\sum_{k'\in A_{c'}}$ $\mathcal{C}_{k,k'}$.
The results $C_{IC}$ in Fig.\ref{fig:correlation} show that inter-class filter correlation in CSG is about half of that in STD, both on AlexNet and ResNet20.
This demonstrates that different classes tend to activate uncorrelated filters in CSG CNNs, aka, CSG CNNs yield highly class-related
	representations where the representations for different classes have less overlap.

\section{Application}\label{sec:app}
Using the class-specificity of filters, we can improve filters' interpretability on object localization. Moreover, the highly class-related representation makes it easier to distinguish abnormal behavior of adversarial samples.

\subsection{Localization}\label{sec:loc}

In this subsection, we conduct experiments to demonstrate that our class-specific filters can localize a class better, for CSG training is demonstrated to encourage each filter in the penultimate layer to focus on fewer classes.

\textbf{Localization method\quad}

Gradient maps~\cite{simonyan2013deep} and activation maps (resized to input size) is a widely used method to determine the area of objects or visual concepts,
which not only works in localization task without bounding box labels~\cite{bau2017network},
but also take an important role in network visualization and understanding the function of filters~\cite{zhou2016learning}.
We study CSG CNNs' performs on localizing object classes with three localization techniques based on filters, including gradient-based saliency map (GradMap)~\cite{simonyan2013deep} and activation map (ActivMap)~\cite{bau2017network} for a single filter and classification activation map (CAM)~\cite{zhou2016learning} for all filters. See Appendix~\ref{app:localization} for the localization techniques.

\begin{table}[t]
    \caption{The performance of localization with resized activation maps in the CSG/STD CNN. For almost all classes, CSG CNN significantly outperforms STD CNN both on Avg-IoU and AP20/AP30.
    }
    \label{tab:local}
	\centering
	\begin{scriptsize}
    \begin{tabular}{l|l|c|c|c|c|c|c|c|c}
    \toprule
    localization  &  metric  & training  & bird & cat & dog & cow & horse & sheep & total \\
    \midrule

    \multirow{4}*{\tabincell{c}{GradMap\\for\\one filter}}  &
	\multirow{2}*{Avg-IoU} & CSG &
	\textbf{0.2765} & \textbf{0.2876} & \textbf{0.3313} & \textbf{0.3261} & \textbf{0.3159} & \textbf{0.2857} & \textbf{0.3035} \\
	~ &  ~ & STD     &
	0.2056 & 0.2568 & 0.2786 & 0.2921 & 0.2779 & 0.2698 & 0.2606 \\
	\cmidrule{2-10}
	~ & \multirow{2}*{AP20} &  CSG &
	\textbf{0.6624} & \textbf{0.8006} & \textbf{0.9170} & \textbf{0.8828} & \textbf{0.8204} & \textbf{0.8089} & \textbf{0.8165} \\
	~ & ~& STD &
	0.4759 & 0.7081 & 0.7556 & 0.8069 & 0.7621 & 0.7764 & 0.7029 \\

	\midrule
    \multirow{4}*{\tabincell{c}{ActivMap\\for\\one filter}} & \multirow{2}*{Avg-IoU}    & CSG &
	     \textbf{0.3266} & \textbf{0.4666} & \textbf{0.4003} & \textbf{0.4226} & \textbf{0.3357} & \textbf{0.4343} & \textbf{0.4289}\\
	    ~ & ~ & STD  &
    	 0.2858 & 0.3816 & 0.3446 & 0.3674 & 0.3019 & 0.3652 & 0.3602 \\
    \cmidrule{2-10}
    ~  & \multirow{2}*{AP30} & CSG &
	    \textbf{0.5867} & \textbf{0.8608} & \textbf{0.7497} & \textbf{0.8072} & \textbf{0.5824} & \textbf{0.8603} & \textbf{0.8251} \\
    ~ & ~ & STD &
    	0.4739 & 0.7004 & 0.6372 & 0.7214 & 0.5157 & 0.6970 & 0.6759 \\

    \midrule
    \multirow{4}*{\tabincell{c}{CAMs\\for\\all filters}}  &  \multirow{2}*{Avg-IoU} & CSG &
    \textbf{0.3489} & \textbf{0.4027} & \textbf{0.3641} & \textbf{0.3972} & \textbf{0.3524} & \textbf{0.3562} & \textbf{0.3694} \\
    ~ &  ~ & STD     &
    0.3458 & 0.3677 & 0.3492 & 0.3516 & 0.3170 & 0.3470 & 0.3483 \\
    \cmidrule{2-10}
    ~ & \multirow{2}*{AP30} &  CSG &
    0.6431 & \textbf{0.8382} & \textbf{0.7197} & \textbf{0.7517} & \textbf{0.7136} & \textbf{0.7073} & \textbf{0.7300} \\
    ~ & ~& STD &
    \textbf{0.6495}& 0.7832 & 0.7085 & 0.6621 & 0.5825 & 0.6504 & 0.6853 \\

    \bottomrule
    \end{tabular}
    \end{scriptsize}
\end{table}

\textbf{Quantitative evaluation\quad}We train ResNet152s to do classification on preprocessed PASCAL VOC and use Avg-IoU (average intersection over union) and AP20/AP30 (average precision 20\%/30\%) to evaluate their localization. Higher metrics indicate better localization. See Appendix~\ref{app:loc_metric} for a detailed definition of the metrics. The results for localization with one or all filters are shown in Table \ref{tab:local}.
For localization with one filter, most classes are localized better with CSG CNN.
That's because CSG encourages filters to be activated by the labeled class rather than many other classes, which alleviates other classes' interference on GradMaps and ActivMaps for each filters.
Furthermore, as a weighted sum of better one-filter activation maps, CSG also outperform STD on CAMs.

\textbf{Discussion} It is widely recognized~\cite{bau2017network} that localization reveals what semantics or classes a filter focuses on. Compared with a vanilla filter, a class-specific filter responds more intensively to the region of relevant semantics and less to the region of irrelevant semantics like background. Thus localization is improved. For STD training, confusion on the penultimate layer is caused by all convolutional layers, however, in our CSG training they are jointly trained with back-propogation to disentangle the penultimate layer.

\begin{figure}[b]
	\centering
    \includegraphics[width=0.9\textwidth]{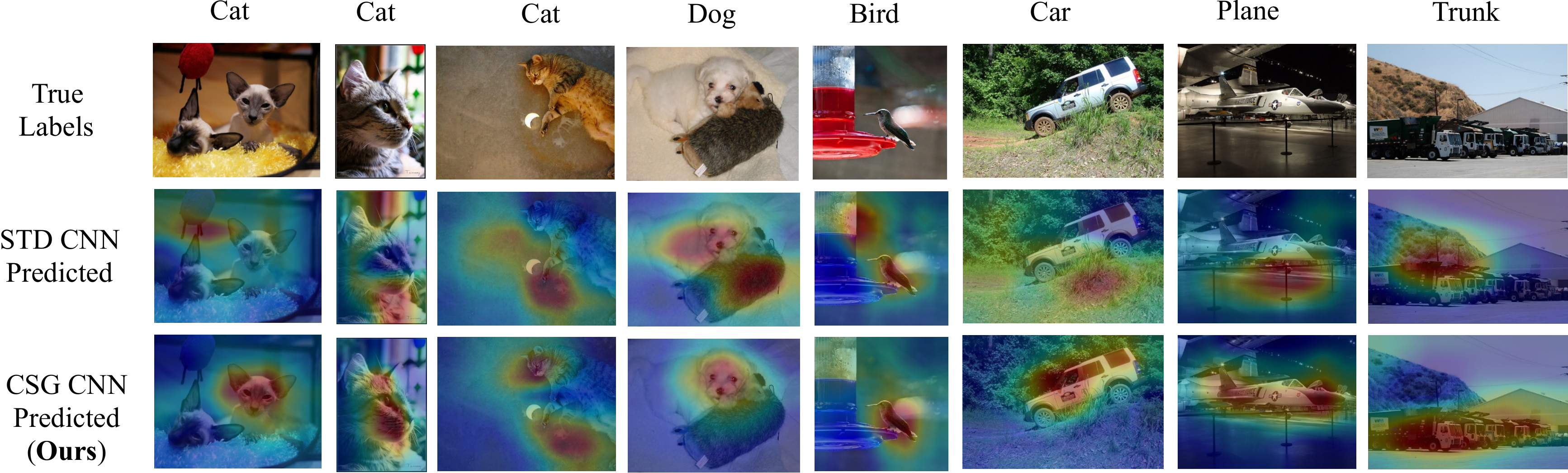}
    \caption{
    Visualizing the localization in STD CNN and CSG CNN with CAM ~\cite{zhou2016learning}.
    }
    \label{fig:cam}
\end{figure}

\textbf{Visualiziation\quad}Besides the quantitative evaluation above, in Fig.\ref{fig:cam}, we also visualize some images and their CAMs from the STD/CSG CNN (ResNet20s trained on CIFAR-20) with ImageNet~\cite{imagenet_cvpr09}. We observe that the CAMs of STD CNN often activate extra area beyond the labeled class. However, CSG training successfully helps the CNN find a more precise area of the labeled objects.
Such a phenomenon vividly demonstrates that CSG training improves the performance in localization.

\subsection{Adversarial Sample Detection}\label{sec:detect}

This subsection shows that the highly class-related representation of our CSG training can promote adversarial samples detection.
It is studied that adversarial samples can be detected based on the abnormal representation across layers~\cite{wang2018interpret}: in low layers of a network, the representation of an adversarial sample is similar to the original class, while on the high layers it is similar to the target class.
Such mismatch is easier to be detected with the class-related representation from CSG, where the implied class are more exposed.

\begin{floatingtable}[r]
	{
		\begin{scriptsize}
	    \begin{tabular}{l|c |c |c|c}
	    \toprule
	    \multicolumn{2}{c|}{Num. of training samples} &  500 & 1000 & 2000\\
	    \midrule
	    \multirow{2}{*}{FGSM} & STD & 4.76 & 4.60 & 4.15  \\
	    & CSG (Ours) & \bf{4.50} & \bf{3.94} & \bf{3.84} \\
	    \hline
	    \multirow{2}{*}{PGD} & STD & 5.03 & 6.20  & 4.08  \\
	    & CSG (Ours) & \bf{4.52} & \bf{3.95} & \bf{3.25} \\
	     \hline
	    \multirow{2}{*}{CW} & STD & 7.33 & 6.76  & 6.46  \\
	    & CSG (Ours) & \bf{7.03} & \bf{6.18} & \bf{5.87} \\
	    \bottomrule
	    \end{tabular}
	    \end{scriptsize}
	}
    \centering
    \caption{The mean error rate(\%) for random forests on adversarial samples detection with the features of CNN.}
    \label{tab:adversarial detection}
\end{floatingtable}

To verify this judgment, we train a random forest ~\cite{breiman2001random} with the features of normal samples and adversarial samples extracted by global average pooling after each convolution layers of ResNet20 trained in \ref{sec:train}.
We generate adversarial samples for random targeted classes by commonly used white-box attacks such as FGSM~\cite{goodfellow2014explaining} , PGD~\cite{madry2017towards}, and CW~\cite{carlini2017towards}.
See Appendix~\ref{app:advdetect} for detailed attack settings.
We repeat each experiment five times and report the mean error rates in Table \ref{tab:adversarial detection}.
The experimental results demonstrate that the class-related representation can better distinguish the abnormal behavior of adversarial samples and hence improve the robustness of the model.
Further experiments in Appendix~\ref{app:robustness} show that CSG training can improve robustness in defending adversarial attack.

\section{Conclusion}

In this work, we propose a simply yet effective structure -- Class-Specific Gate (CSG) to induce filter differentiation in CNNs.
With reasonable assumptions about the behaviors of filters, we derive regularization terms to constrain the form of CSG.
As a result, the sparsity of the gate matrix encourages class-specific filters, and therefore yields sparse and highly class-related representations, which endows model with better interpretability and robustness.
We believe CSG is a promising technique to differentiate filters in CNNs.
Referring to CSG's successful utility and feasibility in the classification problem, as one of our future works, we expect that CSG also has the potential to interpret other tasks like detection, segmentation, etc, and networks more than CNNs.

\textbf{Acknowledgement\quad}This work was supported by the National Key R\&D Program of China (2017YFA0700904), NSFC Projects (61620106010, U19B2034, U1811461, U19A2081, 61673241, 61771273), Beijing NSF Project (L172037), PCL Future Greater-Bay Area Network Facilities for Large-scale Experiments and Applications (LZC0019), Beijing Academy of Artificial Intelligence (BAAI), Tsinghua-Huawei Joint Research Program, a grant from Tsinghua Institute for Guo Qiang, Tiangong Institute for Intelligent Computing, the JP Morgan Faculty Research Program, Microsoft Research Asia, Rejoice Sport Tech. co., LTD and the NVIDIA NVAIL Program with GPU / DGX Acceleration.



\clearpage
\appendix

\section{The Theoretical Convergence Interval for L1-density}
\label{app:l1density}
This section derives the theoretical convergence interval for the L1-density of the CSG matrix $G\in [0,1]^{C\times K}$.
L1-density is defined as $\frac{\left\|G\right\|_1}{CK}$. Now let's find the bound for the L1-density when a CSG CNN converges.

\textbf{Lower bound\quad} In CSG training, we use projected gradient descent (PGD) to restrain $G$ in the solution space
	$\{G\in [0,1]^{C\times K} | \left\|G^k\right\|_\infty = 1 \}$.
	Therefore, for  any $G$ in the space, it is ensured that
	$$\frac{\left\|G\right\|_1}{CK}
	=\frac{ \sum_{k=1}^K \left\|G\right\|_1}{CK}
	\geq \frac{ \sum_{k=1}^K \left\|G\right\|_\infty}{CK}
	= \frac{ \sum_{k=1}^K 1}{CK}
	= \frac{1}{C}. $$
	Therefore, $ \frac{1}{C}$  is lower bound for the L1-density of $G$, which also holds when the CSG CNN converges.

\textbf{Upper bound\quad}  In CSG training, we use  $\lambda_2 d(\left\|G\right\|_1,g)$ as the sparsity regularization to punish $\left\|G\right\|_1$ when it is larger than $g$ which is a hyperparameter as the upper bound for $\left\|G\right\|_1$. If we set $\lambda_2$ as a relatively large number, the sparsity regularization is
	strong enough to reduce $\left\|G\right\|_1$ under $g$ before convergence. Therefore, we get the upper bound for the L1-density of $G$ on convergence as
	$$
	\frac{\left\|G\right\|_1}{CK}
	\leq \frac{g}{CK}.
	$$

Combining the lower bound and the upper bound above, the convergence interval for the L1-density of $G$ is $[\frac{1}{C} ,\frac{g}{CK} ]$.

\section{Training setting and dataset preprocess}\label{app:train}

The ResNet20s are trained on CIFAR-10~\cite{krizhevsky2009learning}. The default settings include: batch size=256; SGD optimizer with momentum=0.9~\cite{sutskever2013importance}; initial learning rate=0.1; total training epochs=200; and every 1 in 3 epochs are in CSG path.

The ResNet18s are trained on ImageNet~\cite{imagenet_cvpr09}. The default setting are the same as ResNet20s except trained on 4 gpus for 120 epochs .

The ResNet152s are finetuned on PASCAL VOC 2010~\cite{pascal-voc-2010} from model pretrained on ImageNet~\cite{imagenet_cvpr09}. Parameters are frozen in and below the second/third resnet layer\footnote{A ResNet152 contains a beginning convolutional layer, 4 resnet layers and a linear layer. Each resnet layer contains a number of residual modules.} for STD/CSG CNN. The first 10 epochs are trained in STD path, and after that every 2 in 3 epochs are in CSG path. The setting is: batch size=32; Adam optimizer~\cite{Adam}; initial learning rate=1e-5 for STD path, 1e-3 for CSG path; total training epochs=150.

We preprocess PASCAL VOC to be a classification dataset for training ResNet152s: we crop out images for the objects in  6 classes (bird, cat, dog, cow, horse and sheep) and resize the image to 128x128; then randomly reassign 3644 objects for training and 1700 objects for testing. No segmentation label is used in training. In testing phase, we only run the STD path which reuses the weights in the CSG path as shown in Fig~\ref{fig:framework}.

The choice of backbone network are meticulously considered:
1) We finetune ImageNet models on a subset of PASCAL with quite few samples. We chose resnet152 to ensure the performance of baselines; 2) training large models on ImageNet from scratch is costly, while resnet18 is also a common choice on ImageNet; 3) We trained resnet20/50/101 on CIFAR10 and they consistently support our CSG conclusion. Since the penultimate layer in resnet20 has only 64 filters which ease visualization in Fig 4 and 7(a1,a2).

\section{Similarity Between Feature Vectors and Gates} \label{app:sim}
To measure the directional similarity between the feature vector for class $y$ and the gate vector for a class $c$,
we design a similarity based on cosine.

For a pair of image and label $(x,y)$ in the dataset $D$, input $x$ into the CSG CNN, we get the average-pooled activation from the class-specific filters. Let's call it the feature vector for $x$ and denote it as $a(x)\in \mathbb{R}^K$.
Therefore the mean feature vector for class $y$ on dataset $D$ is
$$a_y(D) = \underset{(x,y)\in D}{\mathrm{mean}}a(x).$$
Meanwhile, the gate vector for class $c$ is $G_c$, the $c$-th row in the CSG matrix $G$.

Thus we can define the directional similarity between the feature vector for class $y$ and the gate vector for a class $c$ as
$$S_{yc}(D)=\cos(a_y(D), G_c).$$
In this way, we get a similarity matrix $S(D)\in[0,1]^{C\times C}$ for the dataset $D$. If we take $D$ as the set of all true positive samples and all false negative samples, we can calculate the similarity matrices $S_{TP}, S_{FN}$ respectively. Intuitively, a larger directional similarity $S_{yc}(D)$ means the feature vector is more closely related to the classes.

\section{Explanation for Filter Orthogonality }\label{app:orthogonnal}
In this part, we give an intuitive explanation about why CSG training encourages filters for different classes become orthogonal (Paper Sec~\ref{sec:amb}).
Given a class $c$ and a gate matrix that assigns the filter $k$ for class $c$ and filter $k'$ for other class.
During training, filter $k'$ is blocked (i.e., its activation is masked) in the CSG path when class $c$'s images input.
In order to ensure the STD and CSG path generate similar outputs, the filter $k'$ tends to be activated by class $c$ as less as possible, which implies the weight of filter $k'$ is approximately  perpendicular to $V_c$ (the linear space spanned by class $c$'s features in a layer before).
The filter $k$ for class $c$, however, tends to be activated by class $c$ as saliently as possible so as to enable the CNN to recognize this class. So the weight of filter $k$ is approximately within $V_c$.
Overall, the weights of filter $k$ and filter $k'$ tends to be orthogonal.

\section{Manually Fixed Gate Matrix}\label{app:tight}

\begin{figure}[ht]
        \begin{center}
            \includegraphics[width=\textwidth]{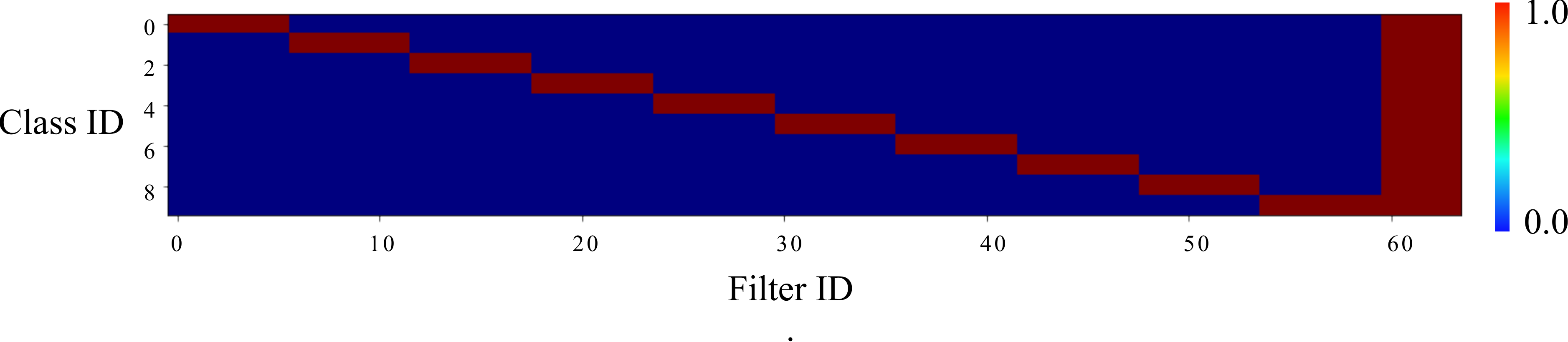}
            \caption{Manually fixed gate matrix for the ResNet on CIFAR10 trained in Paper Sec~\ref{sec:tight}.  Each class monopolizes 6 filters and 4 extra filters are shared by all classes.}
            \label{fig:manual_cd_gate}
        \end{center}
    \end{figure}

In Paper Sec~\ref{sec:tight} we manually initialize the gate matrix and fix it when training ResNet20 on CIFAR-10 from scratch. The gate matrix is visualized in Fig.~\ref{fig:manual_cd_gate}.

We apply this setting based on a statistic analysis on 20 different CSG ResNet20s on CIFAR-10, which aims to figure out how many filters are monopolized by a class.
 The converged CSG matrices indicate that each class tends to monopolize about $m_1=6$ filters and the rest about $m_2=4$ filters are shared by classes.
 Following the statistic analysis, we tighten the constraint by manually setting a fixed CSG matrix for ResNet20s, where each class monopolizes 6 filters and 4 extra filters are shared by all classes.
 Similarly, we set $m_1=25$ and $m_2=6$ for AlexNets.
 They are the CSG matrices we use in Paper Sec~\ref{sec:tight}.

\section{Cluster center experiments}\label{app:cluster}
\begin{figure}[ht]
\begin{center}
	\includegraphics[width=\textwidth]{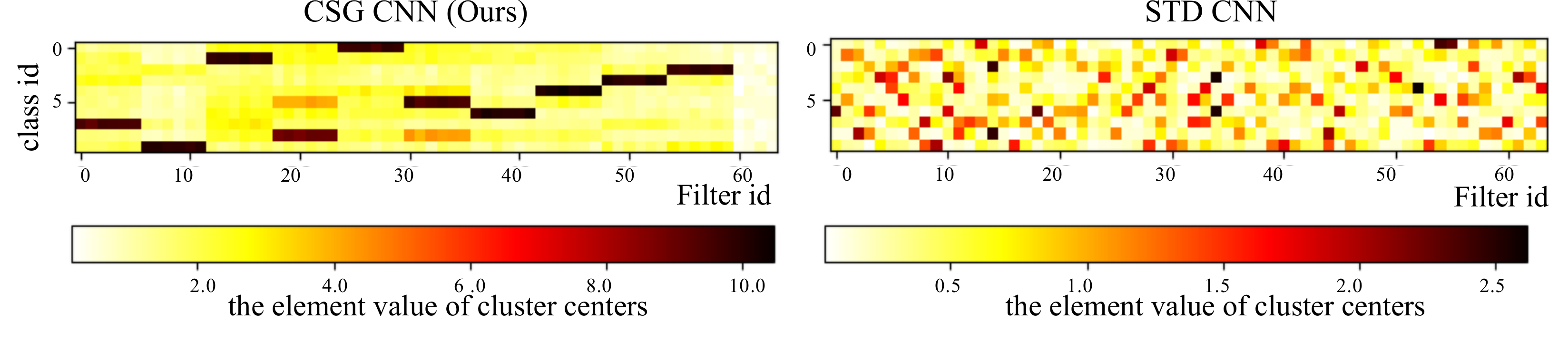}
	\caption{K-means cluster center of the feature vectors from ResNet20 train in Paper Sec~\ref{sec:tight}. The x-axis is the channel id, and the y-axis is class id.  Each row is a the mean of a cluster in the feature vectors' space and the color represents the value of an element in the mean. }
	\label{fig:app-cluster}
\end{center}
\end{figure}

Using the model with fixed gate matirx mentioned in Paper Sec~\ref{sec:tight}, we train ResNet20s on CIFAR-10 with joint CSG and STD training.
Then we run k-means clustering on the feature vectors after the global average pooling in the CSG/STD CNN.
The clustering centers are visualized in Fig.~\ref{fig:app-cluster}.
We find that compared to the STD CNN, the CSG CNN yields better clustering centers, which form groups by channel that is almost the same as the gate matrix visualized in Fig.~\ref{fig:manual_cd_gate}.

\section{Localization Techniques}\label{app:localization}

In Paper Sec~\ref{sec:loc}, we study CSG CNNs' performs on localizing object classes with three localization the techniques based on filters, including gradient-based saliency map (GradMap)  and activation map (ActivMap) for a single filter and classification activation map (CAM) for all filters.

To get a GradMap, we calculate the gradient of a filter's average-polled activation with respect to the input image. Then we normalized the gradient map with its second moment, apply gaussian blur (sigma=5 pixels) and segment out the region with values above $1.0$.

To get an ActMap, we bilinear interpolate the activation map of a filter to input size and segment the region with values above the top 30\% activation of the filter on the entire test dataset.

To get a CAM, we sums up all filters' activation maps with the weights of the linear connections between each channels and an output class~\footnote{CAM only works for CNNs ended with a global average pooling and one linear layer.}.
	By bilinear interpolating the sum activation map to input size
	and segment the region with the top 30\% values in it,
	we get a classification activation map (CAM)~\cite{bau2017network},
	which is segmentation map for a class.

\section{Metrics for Localization} \label{app:loc_metric}
This section gives detailed definition of the metrics we use to evaluate our localization performance in Paper Sec~\ref{sec:loc}.

\subsection{Localization with One Filter}

For an image $x$ in class $c$ (denoted as $x\in D_c\subset D$, where $D$ is the dataset, $D_c$ is the set of images with label $c$), we denote the ground-truth segmentation map for $x$ as $S_x\in\{0,1\}^{H\times W}$, and denote the segmentation map given by filter $k$  as $\hat{S}_x^k\in\{0,1\}^{H\times W}$. The $\hat{S}_x^k$ is calculated as $\hat{S}_x^k=\mathrm{I}\{\mathrm{resize}(A_k)\geq threshold\}$, which means resizing $A_k$ (the activation map from filter $k$) to input size and then thresholding it.

\textbf{The metrics for a filter} on localization is defined below.

The IoU (intersection over union) for filter $k$ on image $x$ is defined as
\footnote{For $A,B\in \{0,1\}^{H\times W}$,\\
 define $A \vee B^{H\times W}$ as $(A \vee B)_{ij}=\max(A_{ij},B_{ij})$;\\
define $A \wedge B^{H\times W}$ as $(A \wedge B)_{ij}=\min(A_{ij},B_{ij})$. }
\footnote{$\left\|\cdot\right\|_0$ is the number of non-zero elements.}
$$\mathrm{IoU}_x^k := \frac{\left\| S_x \wedge \hat{S}_x^k \right\|_0}{\left\| S_x \vee \hat{S}_x^k \right\|_0}.$$

The Avg-IoU (average intersection over union) for filter $k$ on localizing class $c$ is defined as
 $$\mathrm{IoU}_c^k := \underset{x\in D_c}{\mathrm{mean}} \ \mathrm{IoU}_x^k. $$

The APn (average precision $n\%$) for filter $k$ on localizing class $c$ is defined as
 $$\mathrm{APn}_c^k := \underset{x\in D_c}{\mathrm{mean}}\ \mathrm{I} \{\mathrm{IoU}_x^k \geq n\% \}.$$

When $c^* = \arg\max_{c} \mathrm{APn}_{c}^k $, we call filter $k$ is focused on localizing class $c^*$, denoted as $k \in F_{c^*}$, where $F_{c^*}$ is the  set of filters focused on localizing class $c$. Therefore the localization performance for filter $k$ can be evaluated with $\mathrm{IoU}^k := \mathrm{IoU}_{c^*}^k$ and  $\mathrm{APn}^k := \mathrm{APn}_{c^*}^k   $.

\textbf{The metrics averaged for all filters} on localization is defined below based on the aforementioned metrics.

(1) The Avg-IoU and APn for localizing class $c$ as
$$\mathrm{IoU}_c := \mathrm{mean}_{k\in F_{c}} \mathrm{IoU}^k ,$$
 and  $$\mathrm{APn}_c := \mathrm{mean}_{k\in F_{c}} \mathrm{APn}^k .$$

(2) The Avg-IoU and APn for localizing all classes is defined as
$$\mathrm{IoU} := \mathrm{mean}_{k\in\{1,2,...,K\}} \mathrm{IoU}^k  ,$$
and $$\mathrm{APn} := \mathrm{mean}_{k\in\{1,2,...,K\}} \mathrm{APn}^k .$$

\subsection{Localization with All Filters}

For an image $x$ in class $c$ (denoted as $x\in D_c$), we denote the ground-truth segmentation map for $x$ as $S_x\in\{0,1\}^{H\times W}$, and denote the segmentation map given by the classification activation map (CAM)~\cite{bau2017network} as $\hat{S}_x\in\{0,1\}^{H\times W}$. The $\hat{S}_x$ is calculated as $\hat{S}_x=\mathrm{I} \{ \mathrm{resize}(\sum_k W_c^k A_k )\geq threshold\}$, where $A_k$ is the activation map of filter $k$, and $W_c^k$ is the weight of the linear connection between filter $k$ and the logit for class $c$.

The IoU (intersection over union) for CAM on image $x$ is defined as
$$\mathrm{IoU}_x := \frac{\left\| S_x \wedge \hat{S}_x \right\|_0}{\left\| S_x \vee \hat{S}_x \right\|_0}.$$

(1) The metrics for localizing a class is defined below.

The Avg-IoU (average intersection over union) for localizing class $c$ is defined as
 $$\mathrm{IoU}_c := \mathrm{mean}_{x\in D_c} \mathrm{IoU}_x .$$

The APn (average precision $n\%$) for localizing class $c$ is defined as
$$\mathrm{APn}_c := \mathrm{mean}_{x\in D_c} \mathrm{I} \{\mathrm{IoU}_x \geq n\% \}.$$

(2) The metrics for localizing all classes is defined below.
The Avg-IoU for localizing all classes is defined as
 $$\mathrm{IoU} := \mathrm{mean}_{x\in D} \mathrm{IoU}_x  ,$$

The APn for localizing all classes is defined as
$$\mathrm{APn} := \mathrm{mean}_{x\in D} \mathrm{I} \{\mathrm{IoU}_x \geq n\% \}.$$

\section{Detailed Settings in Adversarial Sample Detection}\label{app:advdetect}

In Paper Sec~\ref{sec:detect} we generate the non-targeted adversarial samples with commonly used white-box attack. The setting for them is:
FGDM~\cite{goodfellow2014explaining}  ($\epsilon=0.031$), PGD~\cite{madry2017towards} ($\epsilon=0.031$, $iter = 7$) and CW~\cite{carlini2017towards} ($max\_iterations=100$).  The adversarial target classes are from a random permutation of original classes besides each image's true class. We randomly select 500, 1000, 2000 images per class in CIFAR-10 to form different sizes of training datasets and 100 images per class for the testing.

\section{Defending Adversarial Samples}  \label{app:robustness}

 \begin{table}[ht]
    \caption{Black Box Attack on STD CNN and CSG CNN}
    \label{tab:black box attack}
    \centering
    \begin{small}
    \begin{tabular}{l|c|c|c}
    \toprule
    Attack & Metric & STD CNN  & CSG CNN \\
    \midrule
    No Attack & Accuracy & 88.03\% & 88.85\% \\
    \midrule
    Single Pixel Attack & \multirow{2}*{Attack Success Rates} & 14.00\% & 2.00\% \\
    Local Search Attack & &  15.00\% & 2.00\%  \\
    \bottomrule
    \end{tabular}
    \end{small}
\end{table}

    In this experiments, inspired by using class-specific filters to detect adversarial samples, we further explore CSG CNNs' potential in defending adversarial attacks.
    We use the models (CSG/STD ResNet20) and the dataset (CIFAR10) the same as Paper Sec~\ref{sec:detect}.
    Two black box attacks are conducted, including one pixel attack ~\cite{su2019one} and local search attack  ~\cite{narodytska2016simple}.
    They try to fool models according to the model's predicted probability without access to the models' parameters and architectures.
    From the results shown in Table~\ref{tab:black box attack}, we find both the attacks gain attack success rates on the CSG CNN much lower than on the STD CNN.
    This demonstrates that CSG training also improves robustness of CNNs in defending adversarial attacks.
    We guess the robustness is caused by the increase of within-class distance and the decrease of between-class distance, which requires further verification yet.
    Robustness on defending adversarial attacks is another valuable characteristic of the highly class-related representation from our class-specific filters.

%
%
\bibliographystyle{splncs04}
\bibliography{reference}
\end{document}